\documentclass{article} 
\usepackage{iclr2025_conference,times}

\usepackage{amsmath,amsfonts,bm}

\def\eqref#1{equation~\ref{#1}}

\def\1{\bm{1}}

\DeclareMathAlphabet{\mathsfit}{\encodingdefault}{\sfdefault}{m}{sl}
\SetMathAlphabet{\mathsfit}{bold}{\encodingdefault}{\sfdefault}{bx}{n}

\usepackage{hyperref}
\usepackage{url}
\usepackage{booktabs}       
\usepackage{graphicx}
\usepackage{subcaption}

\title{Complementary, Not Cumulative: Interaction Effects in Physics-Informed Neural Networks for Navier–Stokes Vortex Shedding}

\author{Devesh Shah \\
Independent Researcher \\
\textit{deveshs@umich.edu} \\
}

\iclrfinalcopy 
\begin{document}

\maketitle

\begin{abstract}
Physics-informed neural networks (PINNs) embed governing partial differential equations directly into the training loss, offering a promising alternative to costly CFD solvers for unsteady flows. Yet the growing list of techniques proposed to improve PINN training is typically validated one at a time, leaving open whether these techniques actually compose. We study this question in depth on the DFG/Schäfer–Turek unsteady cylinder wake benchmark. In isolation, nearly every technique performs no better than an untreated baseline. However, combining periodic (SIREN) activations with causal weighting unlocks a previously inaccessible regime, reconstructing velocity and pressure fields to within 4.1\% average relative L2 error against an OpenFOAM reference solution. Adding further techniques instead causes catastrophic performance degradation, demonstrating that individually effective PINN interventions can interact nonlinearly and that more elaborate training recipes are not necessarily better.
\end{abstract}

\section{Introduction}
\label{sec:intro}

Compared to the internet-scale corpora that power today's text and vision based foundation models, most scientific domains are inherently data-starved \citep{willard2022integrating}. Many of the systems being modeled in science, however, are constrained by well-established physical laws. Physics-informed neural networks (PINNs) exploit this by embedding governing equations and boundary conditions directly into the training loss, approximating solutions to partial differential equations (PDEs) with only a fraction of the data such problems would otherwise require \citep{raissi2019physics}.

Since this formulation was introduced, PINNs have been applied across a wide range of domains, including solid mechanics \citep{haghighat2021physics}, heat transfer \citep{cai2021physics}, and biomedical flow modeling \citep{kissas2020machine}. However, these models face well-documented challenges when scaling to more complex problems, largely due to training instability when the loss depends on higher-order derivatives \citep{krishnapriyan2021characterizing, wang2021understanding}.

Fluid dynamics is among the most consequential of these complex problem classes: predicting fluid behavior underpins nearly every discipline of physical engineering. Because traditional computational fluid dynamics (CFD) solvers become especially costly at higher Reynolds numbers \citep{kochkov2021machine}, there is strong motivation for faster learned surrogates such as PINNs. Yet this promise is not automatic: prior work has documented failures even for standard setups \citep{chuang2023predictive}. As PINN research has grown, so has the number of proposed fixes, typically validated in isolation against different baselines. We systematically test whether these techniques compose on the standard DFG/Sch\"afer--Turek unsteady cylinder wake benchmark. Our key contributions are:

\begin{enumerate}
    \item A systematic evaluation of widely-used PINN techniques spanning architectural changes, activation functions, and loss reweighting schemes.
    \item Identification of the specific combination (periodic activations paired with causal weighting) needed to reliably learn vortex shedding.
    \item An analysis showing why further additions to this pairing, such as fourier features or self-adaptive weighting, each independently break it down.
\end{enumerate}

Code for all experiments is publicly available at \url{https://github.com/deveshshah1/Navier_Stokes_Exploration_with_PINNs.git}.
\section{Background \& Related Work}
\label{sec:background}

The Navier–Stokes equations rank among the most notoriously difficult systems in mathematical physics: proving that smooth solutions always exist in three dimensions is one of the seven unsolved Millennium Prize Problems \citep{carlson2006millennium}. These equations describe how a fluid's velocity field evolves under the forces acting on it:

\begin{subequations}\label{eq:navier-stokes}
\begin{align}
    \nabla \cdot \mathbf{u} &= 0 \label{eq:continuity} \\
    \frac{\partial \mathbf{u}}{\partial t} + (\mathbf{u} \cdot \nabla)\mathbf{u} &= -\frac{1}{\rho}\nabla p + \nu \nabla^2 \mathbf{u} \label{eq:momentum}
\end{align}
\end{subequations}

where $\mathbf{u} = (u, v)$ is the velocity field, $p$ is pressure, $\rho$ is fluid density, and $\nu$ is kinematic viscosity. Two properties make this system hard to solve directly: the convective term $(\mathbf{u} \cdot \nabla)\mathbf{u}$ is nonlinear, ruling out closed-form solutions in general, and pressure has no evolution equation of its own, so it must be solved for indirectly at every step to keep the velocity field divergence-free. This coupling is why conventional CFD solves a discretized version of these equations on a spatial mesh, advancing the solution forward in time while iteratively coupling the pressure and velocity fields to enforce incompressibility \citep{patankar2018numerical}. The Reynolds number, the ratio of convective to viscous forces, governs this cost: larger values push flows into unsteady, vortex-dominated regimes (e.g. the unsteady loading on turbine blades) that demand finer meshes and smaller time steps.

PINNs have famously struggled as target dynamics grow more complex. \citet{krishnapriyan2021characterizing} showed that this failure is not due to insufficient network capacity: the same architecture that fails as a PINN can represent the correct solution when trained in a purely supervised fashion. Instead, embedding the PDE residual into the loss creates an optimization landscape that becomes extremely difficult to navigate. \citet{wang2021understanding} identified a related but distinct failure mode: gradients backpropagated from different loss terms (physics, boundary, and initial conditions) can become severely imbalanced during training. Similarly, \citet{rahaman2019spectral} identified a spectral bias in standard MLPs, the baseline PINN architecture: a tendency to preferentially learn low-frequency functions. This bias was later shown to also limit coordinate-based neural representations such as images and 3D scenes \citep{tancik2020fourier}.

Each of these diagnoses led to a wave of new techniques: loss-side fixes such as causal \citep{wang2022respecting} and self-adaptive weighting \citep{mcclenny2020selfadaptive}, and architectural fixes such as Fourier feature encoding \citep{tancik2020fourier} and SIREN activations \citep{sitzmann2020implicit}. These are explored and explained in more detail throughout this paper.

Vortex shedding around a cylinder, the specific problem studied in this work, has also drawn dedicated PINN research in its own right - although published work has largely approached the problem through specialized, one-off loss reformulations or architectural changes. \citet{wei2026bridging} reformulate the loss function itself, adding velocity and pressure correction terms inspired by the classical SIMPLE pressure-velocity coupling algorithm to capture the evolution of vortex shedding. \citet{jiang2025gradient} replace the standard MLP with a U-Net++, predicting a stream function rather than velocity and pressure directly, and approximating derivatives with gradient-free convolutional filters rather than automatic differentiation. \citet{chuang2023predictive} take a more diagnostic approach, using the standard PINN formulation to show that it cannot sustain vortex shedding without continuous data supervision, tracing this failure to numerically dissipative dynamic modes. Our work complements this literature by holding the standard PINN formulation fixed and systematically testing how established, general-purpose techniques combine, rather than proposing a new specialized architecture or loss for this problem.

Beyond PINNs specifically, operator-learning methods such as DeepONet \citep{lu2021learning} and the Fourier Neural Operator \citep{li2020fourier} offer an alternative paradigm for fast CFD surrogates, directly approximating the solution operator of a PDE rather than embedding its residual into the loss. However, unlike our setting, these approaches typically require substantial training data spanning many solved instances; thus, we do not explore these methods further in this work. 
\section{Dataset}
\label{sec:dataset}

We study the two-dimensional flow of an incompressible Newtonian fluid past a circular cylinder confined within a rectangular channel, following the widely adopted Schäfer--Turek (ST) benchmark \citep{schafer1996benchmark}. This benchmark provides well-established reference values for drag and lift coefficients and Strouhal number, making it an ideal testbed for validating PINN approaches to incompressible flow. All simulations use a shared fluid and geometry; only the inlet velocity, and consequently the Reynolds number, differs between cases. We only consider the steady-inlet (ST 2D-1) variant of the benchmark, governed by the Navier--Stokes equations given in \eqref{eq:navier-stokes}.

\begin{figure}[t]
\centering
\includegraphics[width=\textwidth]{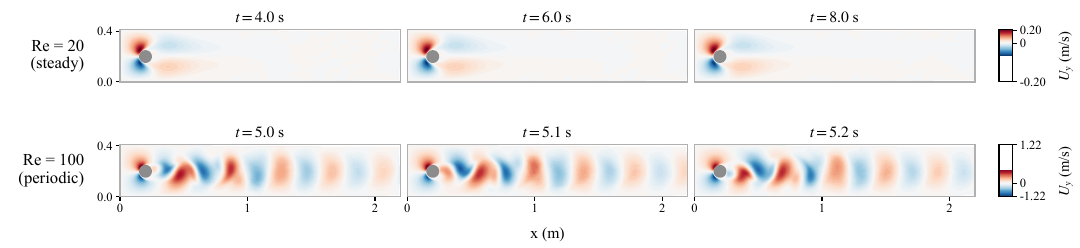}
\caption{$U_y$ velocity field snapshots from our OpenFOAM ground-truth dataset. Top: $Re=20$, steady laminar wake, essentially unchanged across a 4-second span. Bottom: $Re=100$, periodic Kármán vortex street, visibly evolving within a fraction of one shedding period.}
\label{fig:dataset-overview}
\end{figure}

We simulate two Reynolds numbers spanning qualitatively different flow physics, allowing us to quantify how solution complexity affects PINN accuracy. At $Re=20$, the flow is laminar and settles into a steady, symmetric wake. Any temporal variation the PINN predicts after convergence reflects error rather than physics, making this a controlled test of the steady-state limit. At $Re=100$ the flow is unsteady and sheds a periodic Kármán vortex street with a well-defined Strouhal number near 0.30, requiring the network to learn the spatial structure of the shed vortices and their periodic timing. Figure~\ref{fig:dataset-overview} visualizes this contrast directly.

All reference flow fields are generated using OpenFOAM v2512, an open-source finite-volume CFD framework. The mesh is first generated in Gmsh before being imported into OpenFOAM. Both cases are solved with the transient laminar solver \texttt{icoFoam} and PISO pressure-velocity coupling. The mesh is a graded, quad-dominant extrusion with 50{,}184 cells, refined near the cylinder wall and through the wake. For $Re=20$ we simulate $t \in [0,10]$s at $\Delta t = 0.001$s; for $Re=100$ we simulate $t \in [0,8]$s, the duration specified by the benchmark, at $\Delta t = 0.0002$s. Both cases write the full flow field every $0.1$s, yielding 100 and 80 snapshots respectively. Full details appear in Appendix ~\ref{apdx:openfoam}.

Before using these outputs as PINN training and evaluation data, we validate them against the published ST benchmark reference values. At $Re=20$, our converged drag coefficient matches the benchmark to within 0.2\%. At $Re=100$, our time-averaged drag matches to within 7.8\%, while lift amplitude and Strouhal number match to within 1\%. The larger drag discrepancy at $Re=100$ reflects the known sensitivity of integrated drag to near-wall mesh resolution at higher Reynolds number, and caps how closely a PINN trained on these fields can match the true ST solution.

For each simulation, we extract cell-centre velocity, pressure, and coordinate fields from the OpenFOAM output at every written time step. The combined dataset totals approximately 9 million spatiotemporal observations across both Reynolds numbers.
\section{Methods}

\renewenvironment{quote}
  {\list{}{\leftmargin=1em\rightmargin=0pt\parsep=6pt}\item[]\relax}
  {\endlist}

\subsection{Base PINN Formulation}
\label{sec:base-formulation}

We validate the base PINN formulation on the Re=20 case before applying it to the harder Re=100 dynamics studied in the rest of this paper. At Re=20, the flow is steady, so the model only needs to learn a time-invariant, symmetric wake. Training uses no direct data supervision, relying only on governing physics and boundary/initial conditions. At steady state ($t \geq 2.0$s), relative L2 errors against OpenFOAM are all within 3\% (Table~\ref{tab:baseline-comparison}), and the predicted pressure drop across the cylinder matches to within 1\%. This confirms the configuration below is correctly implemented, and this exact recipe is carried over unchanged to Re=100 for the remainder of the paper's ablations.

\begin{quote}
    \textbf{Architecture.} A fully-connected multilayer perceptron maps the spatiotemporal coordinates $(x,y,t)$, normalized to $[-1,1]$ per coordinate using the domain bounds, to the flow variables $(u,v,p)$. The network has 8 hidden layers of width 128, $\tanh$ activations, Xavier-uniform weight initialization, and zero-initialized biases.

    \textbf{Loss formulation.} The training loss is a weighted sum of three terms:
    \begin{equation}
        \mathcal{L} = \lambda_{phys}\mathcal{L}_{phys} + \lambda_{bc}\mathcal{L}_{bc} + \lambda_{ic}\mathcal{L}_{ic},
        \label{eq:total-loss}
    \end{equation}
    with baseline weights $\lambda_{phys}=3.0$, $\lambda_{bc}=5.0$, $\lambda_{ic}=0.2$ and losses defined as: 

    \begin{quote}
        \textbf{$\mathcal{L}_{phys}$} The mean squared residual of the continuity and momentum equations (\eqref{eq:continuity}, \eqref{eq:momentum}), evaluated via automatic differentiation. The momentum residual requires the kinematic viscosity $\nu$, set at $\nu = 0.001$ m$^2$/s for the Re=20 case, consistent with the mean velocity and cylinder diameter in \eqref{eq:reynolds}.

        \textbf{$\mathcal{L}_{bc}$} Sums three boundary terms: the inlet, matched against the analytic Schäfer--Turek parabolic profile given in \eqref{eq:inlet-profile}; the outlet, held at zero pressure; and no-slip, zero velocity on the channel walls and cylinder surface.

        \textbf{$\mathcal{L}_{ic}$} Enforces the fluid starting at rest ($u=v=0$ at $t=0$).
    \end{quote}

    \textbf{Sampling.} Collocation, boundary, and initial-condition points are drawn i.i.d.\ uniformly at random and resampled fresh at every training step: 8000 collocation points, 500 points each at the inlet and outlet, 500 points on each of the three no-slip surfaces, and 300 initial-condition points.

    \textbf{Training.} We train with AdamW, learning rate $10^{-3}$, cosine-annealed over 125K steps, and gradient norm clipped to 1.0.
\end{quote}

\begin{table}[b]
\centering
\caption{Relative L2 error against OpenFOAM for all baseline experiments. $^*$Note Re=20 results are reported at steady state ($t \geq 2.0$s); all other rows report error over the full trajectory.}
\label{tab:baseline-comparison}
\begin{tabular}{cccccc}
\toprule
Re & Physics Loss & Data Supervision & Ux & Uy & p \\
\midrule
20 & \checkmark & $\times$ & 1.1\%$^*$ & 2.1\%$^*$ & 2.8\%$^*$ \\
100 & \checkmark & $\times$ & 68.3\% & 86.2\% & 118.4\% \\
100 & \checkmark & \checkmark & 19.8\% & 70.5\% & 36.1\% \\
100 & $\times$ & \checkmark & 21.6\% & 108.3\% & 35.5\% \\
\bottomrule
\end{tabular}
\end{table}

\subsection{Candidate Techniques}
\label{sec:techniques}

Here we introduce a variety of techniques proposed in the PINN literature, each addressing a specific, previously-documented failure mode of physics-only training (Section~\ref{sec:background}). Each can be added to the base formulation of Section~\ref{sec:base-formulation} independently of the others.

\subsubsection{Fourier Feature Encoding}
\label{sec:ff}
Standard MLPs with smooth activations (e.g. tanh) struggle to learn high-frequency signals: left to gradient descent, they preferentially settle into the smoothest function that satisfies the loss well, a phenomenon known as spectral bias \citep{rahaman2019spectral}. This is a common obstacle in PINN solutions, and especially relevant to ours: vortex shedding is periodic, and with a shedding period of 0.336s within our 8s training window, it creates a genuinely high-frequency signal in time.

Fourier feature encoding counters this by lifting the raw input coordinates into a high-frequency basis before they reach the network \citep{tancik2020fourier}. Each coordinate group is mapped through
\begin{equation}
    \gamma(\mathbf{x}) = [\sin(2\pi B\mathbf{x}),\ \cos(2\pi B\mathbf{x})] \in \mathbb{R}^{2m},
    \label{eq:fourier}
\end{equation}
where the frequency matrix $B$ is sampled once from $\mathcal{N}(0,\sigma^2)$ and then held fixed as a non-trainable buffer for the rest of training. We implement this as two separate encoders, one each for the spatial and time axes, with $\sigma_{spatial}=2.0$ and $\sigma_{time}=1.0$.

\subsubsection{Causal Weighting}
\label{sec:cw}
For time-dependent PDEs, the true solution is causal: the state at time $t$ depends only on the initial condition and the dynamics that unfolded before $t$, not on the solution at later times. Standard PINN training ignores this structure entirely, since collocation points are drawn across the whole time domain. This lets the optimizer reduce the loss by fitting late-time residuals before resolving earlier dynamics, even though a low late-time residual is only physically meaningful if those early dynamics were learned correctly first. \citet{wang2022respecting} argue this is a major source of PINN training failure, and propose weighting the physics loss so that later times only contribute once earlier times are already well fit.

We implement a discretized, batch-local approximation of this idea: each batch's collocation points are binned into $M=20$ equal intervals over $[t_{min}, t_{max}]$, and the mean physics residual within each bin, $\bar{r}_k$, is computed. Bin weights are then set by the cumulative residual in all earlier bins,
\begin{equation}
    w_k = \exp\left(-\epsilon \sum_{j<k} \bar{r}_j\right),
    \label{eq:causal-weight}
\end{equation}
so that a bin only receives substantial weight once the physics residual in every preceding bin has been driven down. Each collocation point inherits the weight of its time bin, and the physics loss becomes the weighted mean $\frac{1}{N}\sum_i w_i r_i$. We set $\epsilon=1.0$.

\subsubsection{Hard Boundary Condition Enforcement}
\label{sec:hbc}
Boundary conditions in a standard PINN are enforced softly, as a term in the loss that the network is only encouraged, not guaranteed, to satisfy. This creates competition with the other loss terms, so the boundary condition can be violated where its weight is outmatched by conflicting physics or data gradients. \citet{sukumar2022exact} propose constructing the network's output so that boundary conditions are satisfied exactly, removing the constraint from the optimization problem.

The no-slip condition on the cylinder surface is a natural candidate for this treatment: it is where the steepest velocity gradients occur, so any imperfection here has outsized downstream effects. We construct a smooth distance-based mask around the cylinder,
\begin{equation}
    d(x,y) = \tanh\left(10 \cdot \frac{r_{dist} - r}{r}\right), \qquad r_{dist} = \sqrt{(x-c_x)^2 + (y-c_y)^2},
    \label{eq:hbc-mask}
\end{equation}
and multiply it into the raw network output for the velocity components, $u = d(x,y)\, u_{raw}$ and $v = d(x,y)\, v_{raw}$. Thus, on the cylinder surface, $d$ vanishes regardless of what the network predicts, enforcing no-slip architecturally rather than through the loss. Away from the cylinder, $d \to 1$, so the mask leaves the solution unaffected outside a thin band near the wall.

\subsubsection{Periodic (SIREN) Activations}
\label{sec:act}
Sinusoidal activations offer a complementary way to address the spectral bias described in Section~\ref{sec:ff}, changing the activation function itself rather than the input encoding. \citet{sitzmann2020implicit} propose replacing $\tanh$ or ReLU with $\sin(\omega z)$ throughout the network, so that every layer, not just the input, is built from oscillatory functions capable of representing fine detail. Since the physics loss requires differentiating the network's output multiple times in the momentum equation, this matters directly for us: repeated differentiation of a sinusoid remains a bounded, well-behaved sinusoid, whereas the derivatives of $\tanh$ degrade and saturate.

Following \citet{sitzmann2020implicit}, we set $\omega$ of the first layer to 30 and 1 for all subsequent layers. SIREN also requires a different initialization scheme: first-layer weights are drawn from $\mathcal{U}(-1/n_{in}, 1/n_{in})$, all later layers from $\mathcal{U}(-\sqrt{6/n_{in}}, \sqrt{6/n_{in}})$, with biases zero-initialized.

\subsubsection{Self-Adaptive Loss Weights}
\label{sec:sa}
Balancing multiple loss terms against each other is a difficult hyperparameter optimization problem. \citet{mcclenny2020selfadaptive} address this by making each weight a learnable parameter and framing the loss as a min-max game: the network minimizes the loss with respect to its weights, while the weights themselves are simultaneously updated to maximize it. This gives training a self-correcting mechanism for loss balancing that responds to how training is actually progressing, rather than a single fixed weighting decided in advance.

We reparameterize each weight in log-space, $\lambda_k = \exp(\ell_k)$, guaranteeing $\lambda_k > 0$ for any value of $\ell_k$. The network weights $\theta$ and the log-weights $\ell$ are optimized separately: $\theta$ is updated by ordinary gradient descent on the loss $L$, while $\ell$ is updated by gradient ascent (learning rate $10^{-3}$) on the same loss:
\begin{equation}
    \ell_k \leftarrow \ell_k + \text{lr} \cdot \nabla_{\ell_k} L = \ell_k + \text{lr} \cdot \lambda_k L_k,
    \label{eq:sa-ascent}
\end{equation}
Since $\nabla_{\ell_k} L = \lambda_k L_k$, this pushes $\lambda_k$ up fastest for whichever component $L_k$ currently has the largest loss. After each update, $\ell_k$ is clamped to $[-3, 5]$ to keep the ascent dynamics from diverging.

\subsubsection{Post-Hoc Second-Order Fine-Tuning (L-BFGS)}
\label{sec:lbfgs}
Adam is a first-order optimizer: it adapts its step size using only the gradient's own recent history, with no information about the curvature of the loss surface around it. This makes it effective at making rapid progress across a rough, high-dimensional loss landscape early in training, but it often fails to converge tightly once training is already close to a local optimum. A common practice in PINN training is to follow Adam with a quasi-Newton method, most often L-BFGS, which builds an approximation of the inverse Hessian from recent gradient history to take curvature-aware steps and refine a solution beyond what Adam alone reaches \citep{raissi2019physics}.

We fine-tune from our best Adam checkpoint using an initial step size of 1.0, rescaled at each step by a strong-Wolfe line search, and an inverse-Hessian approximation built from a history of 100 past gradient-step pairs. Unlike the continuous resampling used during Adam training, the collocation, boundary, initial-condition, and data points are held fixed for each L-BFGS step, so that the loss and gradient stay consistent across the multiple re-evaluations required by the line search.

\subsection{Experimental Design}
\label{sec:experiments}
Applying the base formulation from Section~\ref{sec:base-formulation} to Re=100 fails badly (Table~\ref{tab:baseline-comparison}): relative L2 errors reach 68\% (Ux), 86\% (Uy), and 118\% (p). This is not a noisy version of the correct dynamics, but a collapse to a smoothed, quasi-steady near-field response around the cylinder. The model effectively finds it easier to zero out the temporal derivative than to resolve the true periodic dynamics.

To address this, we test the full set of candidate techniques from Section~\ref{sec:techniques} individually, but find that none meaningfully move the model out of this collapse. We therefore introduce an additional loss term, $\mathcal{L}_{data}$, that grounds training with direct supervision from a held-out 1\% (40,000-point) sample of our OpenFOAM ground truth. Adding this term partially rescues the baseline on Ux and p, but Uy, the velocity component most directly driven by shedding, barely moves (Table~\ref{tab:baseline-comparison}).

To verify the physics loss itself contributes, we test a data-only control with the physics loss removed. It does not outperform the physics-only baseline (Table~\ref{tab:baseline-comparison}), but fails differently: it produces roughly correct oscillation amplitude decorrelated from the true shedding phase, rather than the near-zero variance of the physics-only runs. This confirms physics and data address different parts of the failure, and neither alone suffices. We adopt the physics-constrained, data-supervised configuration as the baseline for all results below; consistent with \citet{chuang2023predictive}, some form of data supervision proves necessary for learning this problem at all.

Given a limited compute budget of T4 GPUs, the base formulation in Section~\ref{sec:base-formulation} is itself intentionally kept at a reduced scale. None of our runs are trained to full convergence and loss continues to decrease slowly well beyond our training budget. We rely on the assumption that relative comparisons between configurations remain informative under this shared, fixed compute budget.

With this data-supervised configuration as our baseline, we evaluate the six candidate techniques from Section~\ref{sec:techniques} in two stages. We first test each technique individually to measure how much of the shedding failure mode any single technique can address on its own. We then use these individual results to guide a series of combination searches, pairing and stacking techniques to see which combinations succeed where the individual techniques fall short, and which instead degrade performance relative to their parts.

\subsection{Evaluation Metrics}
\label{sec:metrics}
To avoid a single aggregate error hiding the qualitatively different failure modes discussed in Section~\ref{sec:experiments}, we report three tiers of metrics. Our primary accuracy measure is the relative L2 error, computed per field over the full spatiotemporal domain. To separate models that fail to oscillate at all from those that oscillate with the wrong phase, we additionally report temporal correlation and oscillation amplitude ratio at fixed downstream wake probes. Finally, we report two domain-standard quantities from the CFD literature, pressure drop across the cylinder and Strouhal number.
\section{Results}

Section~\ref{sec:experiments} established a working baseline. We now turn to this paper's central questions: how much each individual technique matters, which combination reliably learns the shedding dynamics, and, for the combinations that fail, what specific mechanism is responsible.

Table~\ref{tab:full-results} summarizes every individual technique and combination tested. Individually, none of the five techniques recovers any coherent shedding tone or meaningful wake oscillation: L2 error, pressure, and dominant frequency all barely move from baseline. Pairing causal weighting with either spectral-bias fix, SIREN or Fourier Features, changes this qualitatively: both recover the true Strouhal number and produce strong phase-locked correlation at every wake probe, with SIREN the stronger pairing on every metric. Stacking further onto this pairing is where the interesting failures live: tightening causal weighting or doubling data supervision leaves headline error roughly unchanged while modestly improving temporal fidelity, adding Self-Adaptive Weights or Fourier Features on top of a working recipe both actively destroy it: one an optimization failure, the other an architectural one, each traced to a specific mechanism below.

\begin{table}[t]
\centering
\caption{Relative L2 error, predicted Strouhal number, wake-probe amplitude ratio, and wake-probe correlation (each averaged over $x/D=2,4,6$) for various experiments. True St=0.300 for all rows.}
\label{tab:full-results}
\resizebox{\textwidth}{!}{%
\small
\begin{tabular}{lcccccc}
\toprule
Configuration & Ux & Uy & p & St pred & Amp.\ ratio & Corr. \\
\midrule
Baseline & 19.8\% & 70.5\% & 36.1\% & 0.013 & 0.005 & 0.005 \\
\midrule
\multicolumn{7}{l}{\textit{Individual techniques}} \\
\midrule
Fourier Features & 19.4\% & 70.5\% & 35.2\% & 0.025 & 0.007 & 0.003 \\
Causal Weighting & 19.2\% & 70.5\% & 35.4\% & 0.013 & 0.007 & 0.005 \\
SIREN Activation & 19.5\% & 70.5\% & 35.3\% & 0.025 & 0.010 & 0.007 \\
Hard Boundary Conditions & 22.2\% & 72.4\% & 45.0\% & 0.025 & 0.005 & $-$0.002 \\
Self-Adaptive Weights & 36.1\% & 89.8\% & 91.6\% & 0.013 & 0.009 & $-$0.005 \\
\midrule
\multicolumn{7}{l}{\textit{Combinations}} \\
\midrule
Fourier Features + Causal Weighting & 11.7\% & 38.4\% & 21.9\% & \textbf{0.300} & 0.720 & 0.861 \\
Fourier Features + Causal Weighting (2\% data) & 12.1\% & 38.1\% & 22.2\% & \textbf{0.300} & 0.755 & 0.865 \\
SIREN + Causal Weighting & \textbf{7.8\%} & \textbf{17.1\%} & \textbf{16.7\%} & \textbf{0.300} & 0.833 & \textbf{0.988} \\
SIREN + Causal Weighting ($\epsilon=5$) & 9.6\% & 24.5\% & 19.0\% & \textbf{0.300} & \textbf{0.892} & 0.964 \\
SIREN + Causal Weighting + Self-Adaptive & 21.8\% & 75.7\% & 37.9\% & 0.088 & 0.110 & 0.005 \\
Fourier Features + SIREN + Causal Weighting & 58.1\% & 99.8\% & 98.1\% & 0.025 & 0.002 & 0.053 \\
\bottomrule
\end{tabular}%
}
\end{table}

Table~\ref{tab:scaling-results} tests whether SIREN+causal weighting's remaining error is a capacity or training-budget limitation, given the reduced-scale setup of Section~\ref{sec:experiments}: training our current architecture for longer (4000 epochs), training a larger model (6 hidden layers of width 256), and fine-tuning the converged larger model with L-BFGS. Both training longer and using a larger model help substantially, more than halving error, while L-BFGS destroys nearly all of these gains.

\begin{table}[h]
\centering
\caption{Understanding capacity and training budget limitations of the best configuration.}
\label{tab:scaling-results}
\resizebox{\textwidth}{!}{%
\small
\begin{tabular}{lcccccc}
\toprule
Configuration & Ux & Uy & p & St pred & Amp.\ ratio & Corr. \\
\midrule
SIREN + Causal Weighting (reference) & 7.8\% & 17.1\% & 16.7\% & \textbf{0.300} & 0.833 & 0.988 \\
+ more epochs, trained to convergence & 4.4\% & 10.8\% & 10.6\% & \textbf{0.300} & 0.901 & 0.996 \\
+ deeper model, trained to convergence & \textbf{3.0\%} & \textbf{4.8\%} & \textbf{4.6\%} & \textbf{0.300} & \textbf{0.983} & \textbf{1.000} \\
+ L-BFGS fine-tuned (on deeper model) & 22.4\% & 80.3\% & 43.0\% & 0.013 & 0.041 & 0.003 \\
\bottomrule
\end{tabular}%
}
\end{table}

\begin{figure}[t]
\centering
\includegraphics[width=\textwidth]{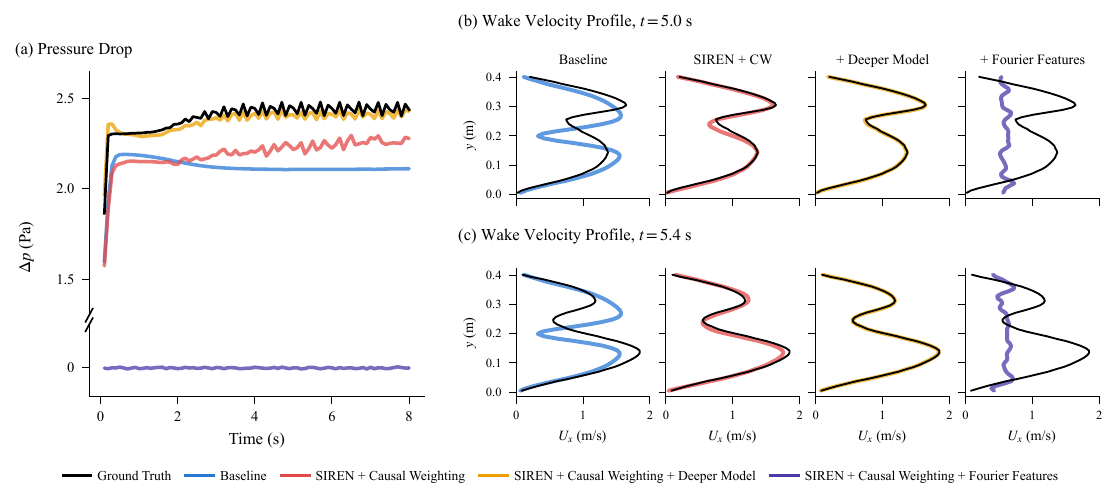}
\caption{For various configurations we compare (a) Pressure drop $\Delta p(t)$ and (b, c) streamwise wake velocity profiles $U_x(y)$ at $t=5.0$s and $t=5.4$s against the ground truth.}
\label{fig:deltap-wake}
\end{figure}

Figure~\ref{fig:deltap-wake} shows the impact of these configurations directly. The baseline never learns the oscillating pressure-drop behavior at all, SIREN+causal weighting recovers it but at reduced amplitude, and the deeper, fully-converged model is nearly indistinguishable from OpenFOAM, including the wake-deficit double-hump structure in the velocity profiles. Fourier Features stacked on SIREN+causal weighting instead collapses entirely, a failure mode we return to in Section~\ref{sec:arch-incompat}.

\subsection{Optimization Instability}
We now examine why these failures occur, starting with a pattern visible directly in the training checkpoints: across many of the experiments, especially the individual techniques tested without pairing, each catastrophic run's best checkpoint was captured within the first 50 epochs of a 2000-epoch schedule. This is a real taxonomy: optimization-dynamics failures break training itself, early and permanently, distinct from configurations that train normally but simply converge to a worse solution.

\begin{figure}[t]
\centering
\includegraphics[width=\textwidth]{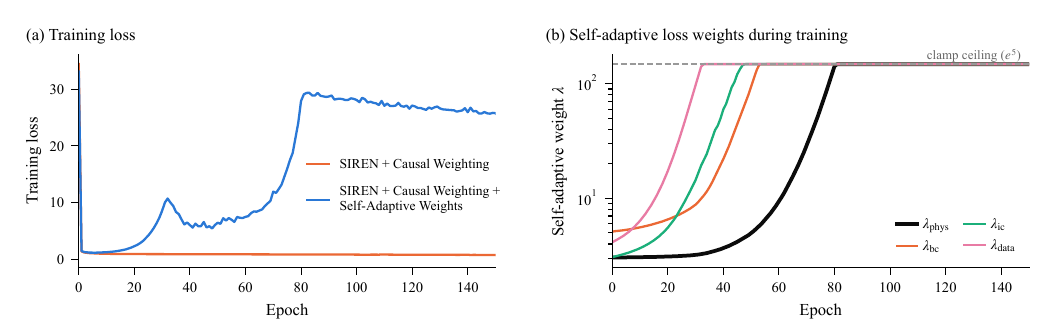}
\caption{(a) Training loss for SIREN+causal weighting with and without Self-Adaptive Weights. (b) Self-adaptive loss weights over training: all four escalate to the clamp ceiling in a staggered order, data first, physics last.}
\label{fig:sa-lambdas}
\end{figure}

We take a closer look at one technique in particular to understand why training freezes so early: Self-Adaptive Weights. Figure~\ref{fig:sa-lambdas}(a) shows the training loss spiking early once Self-Adaptive Weights are introduced. Panel (b) shows why: the four self-adaptive weights escalate toward their clamp ceiling ($e^5$, \eqref{eq:sa-ascent}) in a staggered order, with $\lambda_{phys}$ last. The adversarial ascent inflates whichever term is currently easiest to increase loss on first, boundary and data fitting, well before it reaches the physics residual. By the time every weight has been driven to its bound, the optimization has already been destabilized for the better part of one hundred epochs. This staggered blowup, not a permanently starved physics term, is what produces the early freeze in the best-checkpoint epoch.

L-BFGS's collapse, on top of the converged deeper model, shares this same early-freeze signature but for a distinct mechanism. L-BFGS did not drift or diverge gradually: it took one large step that flattened and redistributed the weights, erasing the sharp, high-magnitude weights SIREN relies on to represent high-frequency content, and its line search then stalled completely. The failure is a single catastrophic update, not a gradual overshoot.

\subsection{Architectural Incompatibility}
\label{sec:arch-incompat}
The collapse of Fourier Features + SIREN + Causal Weighting has a different cause entirely. We compare per-term training loss at initialization (epoch 0) for SIREN+Causal Weighting with and without Fourier Features. Adding Fourier Features inflates the physics residual by $79\times$ relative to SIREN alone, while the other three terms ($\mathcal{L}_{bc}$, $\mathcal{L}_{ic}$, $\mathcal{L}_{data}$) stay within the same order of magnitude. This imbalance is fatal: by epoch 15, the physics loss has collapsed to approximately $4\times10^{-6}$, essentially zero, and stays pinned there for the remaining $\sim$1300 epochs of the run.

\begin{figure}[t]
\centering
\includegraphics[width=\textwidth]{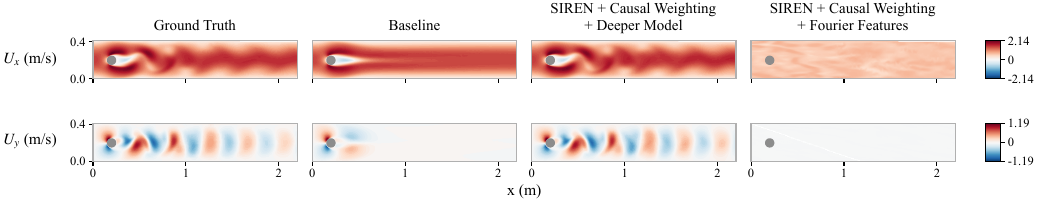}
\caption{Field comparison at $t=5.0$s for various experimental configurations. Note that SIREN + Fourier Features together collapse to a near-uniform field.}
\label{fig:field-comparison}
\end{figure}

The physics residual (\eqref{eq:momentum}) is dominated by derivative terms, $u_x, u_y, u_t, u_{xx}, u_{yy},$ $v_x, v_y, v_{xx}, v_{yy}$, all of which vanish identically if the network's output is constant. Driving physics loss to zero therefore does not require learning the flow field: collapsing the output to a near-constant function trivially satisfies it. Comparing the trained checkpoints' output-layer weights confirms this: the final linear layer shrinks by roughly $21\times$ when Fourier Features is added (output layer norm shrinks from 1.397 to 0.066). This is the direct cause of the field collapse visible in Figure~\ref{fig:field-comparison}, where predicted $U_y$ and pressure both flatten to a narrow band near zero across the whole domain.

Why does stacking Fourier Features onto SIREN specifically cause this, rather than simply adding capacity? The two techniques compound multiplicatively rather than additively at the first layer. A Fourier feature is $\gamma(x)=\sin(2\pi\beta x)$ with $\beta\sim\mathcal{N}(0,\sigma^2)$; feeding it into a SIREN first layer, $h(x)=\sin(\omega_0(w\cdot\gamma(x)+b))$, gives by the chain rule

\begin{equation}
    \frac{dh}{dx} = \omega_0\cos(\cdot)\cdot w \cdot \underbrace{2\pi\beta\cos(2\pi\beta x)}_{d\gamma/dx},
    \label{eq:ff-siren-chain-rule}
\end{equation}
so the effective first-layer frequency is $\omega_0\cdot2\pi\beta$ rather than just $\omega_0$. With our hyperparameters ($\omega_0=30$, $\sigma_{spatial}=2.0$), this is roughly a $2\pi\sigma\approx12.6\times$ amplification, and because the viscous term needs a second derivative, this compounds further. Measuring PDE-residual derivatives at random initialization confirms this structurally: the second-derivative standard deviation is roughly $2.4\times$ larger with Fourier Features added. As a result, compounding frequencies inflate the initial physics residual by roughly $79\times$, and gradient descent takes the cheapest available exit by collapsing the output layer toward a degenerate near-constant solution where every derivative term vanishes trivially.
\section{Conclusion}
\label{sec:conclusion}
We presented a systematic ablation study of five widely-used PINN techniques on the DFG/Schäfer--Turek unsteady cylinder wake benchmark, evaluating each individually, in combination, and stacked beyond their best-performing pairing. This paper's central finding is that these techniques are complementary, not cumulative: which pairing works matters far more than how many techniques are stacked together.

An open question is whether this complementarity generalizes beyond our controlled benchmark, to three-dimensional, higher-Reynolds-number flows where the underlying dynamics are far more complex, and where techniques validated here may compose differently, or not at all. More broadly, we hope the diagnostic approach taken here, tracing failures to specific, checkable mechanisms in training checkpoints and network weights rather than treating them as unexplained instability, proves useful for evaluating other technique combinations beyond the ones studied here.

\bibliography{iclr2025_conference}
\bibliographystyle{iclr2025_conference}

\newpage
\appendix
\section{CFD: OpenFOAM Configuration}
\label{apdx:openfoam}

This appendix documents the full OpenFOAM configuration used to generate the reference dataset in this work: the problem geometry and governing relations, mesh generation, boundary conditions, solver configuration, and the specific parameters used for each of the two Reynolds number cases.

\subsection{Problem Geometry and Governing Relations}

We simulate two-dimensional, incompressible flow of a Newtonian fluid past a circular cylinder confined within a rectangular channel, following the Schäfer--Turek (ST) benchmark \citep{schafer1996benchmark}. The domain is a channel of length $L$ and height $H$, with a cylinder of diameter $D$ positioned off-centre near the inlet (Table~\ref{tab:geometry}). The flow is governed by the incompressible Navier--Stokes equations given in ~\eqref{eq:navier-stokes}, where $\mathbf{u}=(u,v)$ is the velocity field, $p$ is pressure, $\rho$ is fluid density, and $\nu$ is kinematic viscosity. The Reynolds number is defined as
\begin{equation}
    Re = \frac{U_{mean} D}{\nu},
    \label{eq:reynolds}
\end{equation}
where $U_{mean}$ is the mean inlet velocity; it is the only quantity that differs between our two simulated cases (Sections~\ref{app:re20} and \ref{app:re100}). The inlet boundary follows the parabolic (Hagen--Poiseuille) profile specified by the ST benchmark,
\begin{align}
    u(0, y, t) &= \frac{6 U_{mean} y (H-y)}{H^2} \label{eq:inlet-profile} \\
    v(0, y, t) &= 0,
\end{align}
and all simulations are initialised from rest ($u=v=0$ everywhere at $t=0$). Drag and lift coefficients, used throughout this appendix to validate our simulations against the published ST reference values, are computed from OpenFOAM's \texttt{forceCoeffs} function object using reference velocity $U_{ref}=U_{mean}$, reference length $L_{ref}=D$, and reference area $A_{ref} = D \times \text{depth}$:
\begin{align}
    C_d &= \frac{2 F_d}{\rho U_{mean}^2 D \, \text{depth}} \\
    C_l &= \frac{2 F_l}{\rho U_{mean}^2 D \, \text{depth}}
\end{align}

\begin{table}[h]
\centering
\caption{Computational geometry. All dimensions are in metres.}
\label{tab:geometry}
\begin{tabular}{lcc}
\toprule
Parameter & Value & Units \\
\midrule
Channel length $L$ & 2.2 & m \\
Channel height $H$ & 0.41 & m \\
Cylinder centre $(c_x, c_y)$ & (0.20, 0.20) & m \\
Cylinder radius $R$ & 0.05 & m \\
Cylinder diameter $D$ & 0.10 & m \\
Domain extent ($z$) & 0.01 & m (pseudo-2D) \\
\bottomrule
\end{tabular}
\end{table}

\subsection{Mesh Generation and Post-Processing}

The mesh is generated in Gmsh as a two-dimensional quad-dominant surface mesh with a single-layer hexahedral extrusion in $z$, giving a fully structured pseudo-2D domain, and is imported into OpenFOAM with \texttt{gmshToFoam}. Graded refinement toward the cylinder wall and through the wake is achieved with Gmsh distance and threshold fields applied to the cylinder arc curves (Table~\ref{tab:mesh}). The front and back faces of the domain are treated as \texttt{empty} (pseudo-2D) patches.

For each written snapshot, velocity components $u$, $v$, and kinematic pressure $p/\rho$ are read from the OpenFOAM binary-format field files; only internal cell values are extracted, and boundary face values are discarded. Cell-centre coordinates are obtained from the \texttt{C} field written by \texttt{writeCellCentres}, and the $z$-coordinate is discarded since the domain is pseudo-2D.

\begin{table}[h]
\centering
\caption{Mesh generation parameters. The mesh is shared by both $Re=20$ and $Re=100$ simulations.}
\label{tab:mesh}
\begin{tabular}{lcc}
\toprule
Parameter & Value & Units / notes \\
\midrule
Meshing tool & Gmsh 4.x & --- \\
Element type & Hexahedral (quad-extruded) & --- \\
Total cells & 50{,}184 & --- \\
Far-field element size $l_{c,far}$ & 0.030 & m \\
Wake refinement size $l_{c,wake}$ & 0.003 & m \\
Near-cylinder size $l_{c,near}$ & 0.001 & m \\
Extrusion layers ($z$) & 1 & --- \\
Max aspect ratio & 28.9 & --- \\
Max non-orthogonality & 47.2 & degrees \\
Max skewness & 0.94 & --- \\
\bottomrule
\end{tabular}
\end{table}

\subsection{Boundary Conditions}

Table~\ref{tab:bcs} lists the boundary condition applied to each patch. Conditions are identical across both Reynolds numbers; only the inlet velocity magnitude, via $U_{mean}$ in ~\eqref{eq:inlet-profile}, differs between cases.

\begin{table}[h]
\centering
\caption{Boundary condition specification for all patches.}
\label{tab:bcs}
\begin{tabular}{llll}
\toprule
Boundary & Patch type & Velocity ($U$) & Pressure ($p$) \\
\midrule
Inlet ($x=0$) & patch & Parabolic profile (~\eqref{eq:inlet-profile}) & zeroGradient \\
Outlet ($x=2.2$) & patch & zeroGradient & fixedValue $=0$ \\
Top/bottom walls & wall & noSlip & zeroGradient \\
Cylinder surface & wall & noSlip & zeroGradient \\
Front/back faces & empty & empty (2-D) & empty (2-D) \\
\bottomrule
\end{tabular}
\end{table}

\subsection{OpenFOAM Solver Configuration}

Both Reynolds number cases are solved in OpenFOAM v2512 using \texttt{icoFoam}, a transient solver for incompressible, laminar flow; no turbulence model is applied, which is appropriate for $Re<200$. Pressure--velocity coupling is handled by the PISO algorithm, with the correction and scheme settings summarised in Table~\ref{tab:solver}.

\begin{table}[h]
\centering
\caption{Solver and numerical scheme configuration applied to both simulations.}
\label{tab:solver}
\begin{tabular}{ll}
\toprule
Parameter & Value \\
\midrule
Solver & \texttt{icoFoam} (transient laminar NS) \\
Time scheme & Euler (1st-order) \\
Divergence scheme & Gauss linearUpwind \\
Laplacian scheme & Gauss linear corrected \\
Pressure solver & GAMG + Gauss--Seidel \\
Velocity solver & smoothSolver + Gauss--Seidel \\
PISO correctors & 3 \\
Non-orthogonality correctors & 2 \\
Turbulence model & Laminar (none), appropriate for $Re<200$ \\
Fluid density $\rho$ & 1.0 kg/m$^3$ \\
Kinematic viscosity $\nu$ & 0.001 m$^2$/s \\
\bottomrule
\end{tabular}
\end{table}

\subsection{Reynolds Number 20: Steady Flow}
\label{app:re20}

At $Re=20$, obtained with a mean inlet velocity $U_{mean}=0.20$ m/s (~\eqref{eq:reynolds}), the flow is laminar and converges to a steady, symmetric wake by approximately $t=5$s. We simulate to $t=10$s at $\Delta t=0.001$s, well within the stability limit (max Courant number 0.58), and write the full flow field every 0.1s for 100 snapshots. Our converged drag coefficient of $C_d=5.567$ matches the published ST reference value of 5.579 to within 0.2\%, validating the simulation before use as training data (Table~\ref{tab:re20}).

\begin{table}[h]
\centering
\caption{Time integration and output parameters for the $Re=20$ (steady) simulation.}
\label{tab:re20}
\begin{tabular}{lcc}
\toprule
Parameter & Value & Units / notes \\
\midrule
Reynolds number $Re$ & 20 & --- \\
Peak inlet velocity $U_m$ & 0.30 & m/s \\
Mean inlet velocity $U_{mean}$ & 0.20 & m/s $= \tfrac{2}{3} U_m$ \\
Time step $\Delta t$ & 0.001 & s \\
Simulation end time & 10.0 & s \\
Write interval & 0.1 & s (every 100 steps) \\
Number of snapshots & 100 & --- \\
Max Courant number & 0.58 & --- \\
Flow regime & Steady-state & converges $\sim t=5$s \\
$C_d$ (converged) & 5.567 & ST ref.\ 5.579 (0.2\% error) \\
Total data points & 5{,}018{,}400 & 100 snaps $\times$ 50{,}184 cells \\
\bottomrule
\end{tabular}
\end{table}

\subsection{Reynolds Number 100: Unsteady Vortex Shedding}
\label{app:re100}

At $Re=100$, obtained with a mean inlet velocity $U_{mean}=1.00$ m/s, the flow is unsteady and sheds a periodic Kármán vortex street from approximately $t=4$s onward. The higher velocity and finer near-wall dynamics require a smaller time step: we simulate to $t=8$s, the duration specified by the ST benchmark, at $\Delta t=0.0002$s (max Courant number 0.85), again writing the full flow field every 0.1s, for 80 snapshots. Our time-averaged drag coefficient over the periodic regime ($t \in [4,8]$s) matches the ST reference to within 7.8\%, while lift amplitude and Strouhal number both match to within 1\% (Table~\ref{tab:re100}). The larger drag discrepancy reflects the known sensitivity of integrated drag to near-wall mesh resolution at higher Reynolds number; it sets a practical upper bound on how closely a PINN trained on these fields can match the true ST solution, but does not affect the self-consistency of our evaluation framework.

\begin{table}[h]
\centering
\caption{Time integration and output parameters for the $Re=100$ (unsteady) simulation.}
\label{tab:re100}
\begin{tabular}{lcc}
\toprule
Parameter & Value & Units / notes \\
\midrule
Reynolds number $Re$ & 100 & --- \\
Peak inlet velocity $U_m$ & 1.50 & m/s \\
Mean inlet velocity $U_{mean}$ & 1.00 & m/s $= \tfrac{2}{3} U_m$ \\
Time step $\Delta t$ & 0.0002 & s \\
Simulation end time & 8.0 & s (ST benchmark spec) \\
Write interval & 0.1 & s (every 500 steps) \\
Number of snapshots & 80 & --- \\
Max Courant number & 0.85 & --- \\
Flow regime & Unsteady (vortex shedding) & periodic from $\sim t=4$s \\
Mean $C_d$ & 3.170 & ST ref.\ 3.44 (7.8\% error) \\
Max $|C_l|$ & 0.935 & ST ref.\ 0.928 (0.75\% error) \\
Strouhal number $St$ & 0.298 & ST ref.\ 0.300 (0.7\% error) \\
Vortex shedding period $T$ & $\sim$0.336 & s \\
Total data points & 4{,}014{,}720 & 80 snaps $\times$ 50{,}184 cells \\
\bottomrule
\end{tabular}
\end{table}
\newpage
\section{Full Field Comparisons}
\label{apdx:fields}
Figure~\ref{fig:field-comparison} in the main text shows a single-quantity snapshot across four configurations. Here we provide the complete picture for three representative configurations, the baseline, our best-performing pairing scaled to a deeper model, and the Fourier Features + SIREN failure case, showing the true field (OpenFOAM), the PINN prediction, and the absolute error for all three physical quantities ($U_x$, $U_y$, $p$) at $t=5.0$s and $t=5.4$s.
\begin{figure}[!h]
\centering
\includegraphics[width=0.99\textwidth]{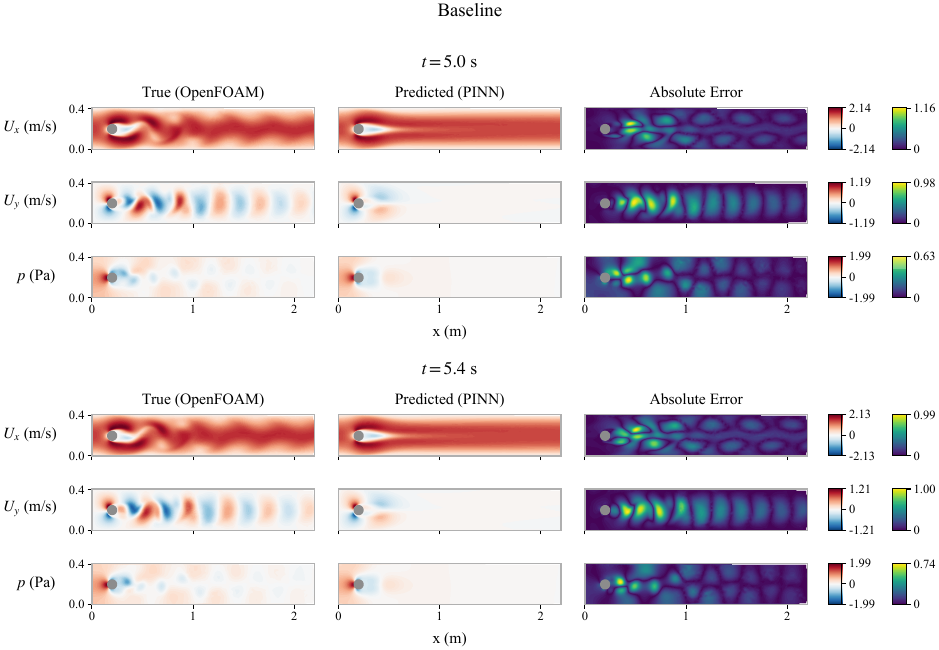}
\caption{Full field comparison for the baseline configuration: true (OpenFOAM), predicted (PINN), and absolute error for $U_x$, $U_y$, and $p$ at $t=5.0$s and $t=5.4$s. The baseline fails to reconstruct any wake oscillation, collapsing to a smoothed, near-steady near-field response.}
\label{fig:apdx-baseline}
\end{figure}

\begin{figure}[p]
\centering
\begin{subfigure}{\textwidth}
\centering
\includegraphics[width=0.99\textwidth]{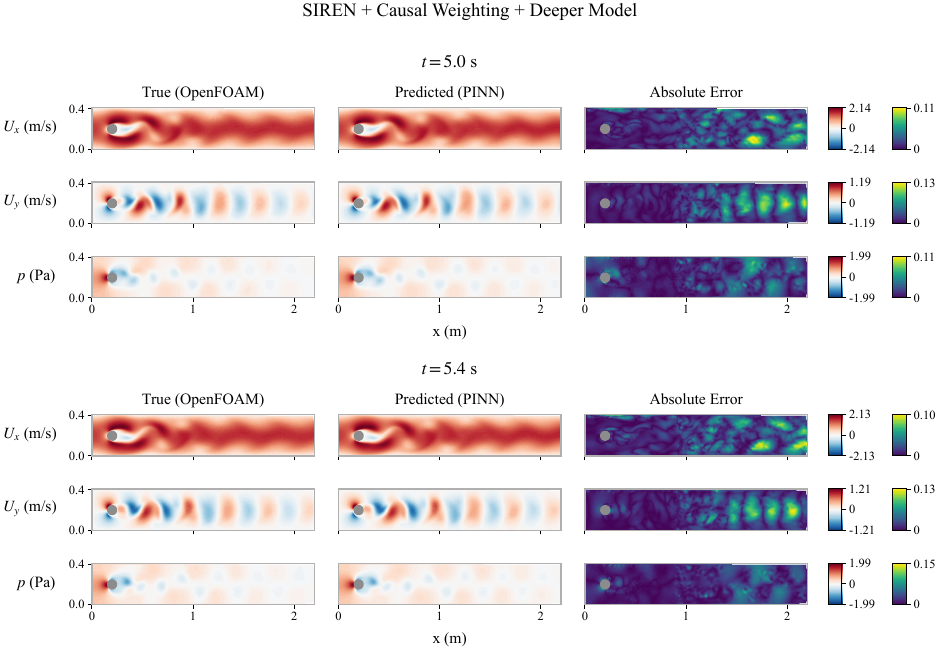}
\caption{SIREN + Causal Weighting + Deeper Model. This configuration reconstructs the full vortex street with error concentrated in the far wake.}
\label{fig:apdx-deeper}
\end{subfigure}

\vspace{3.0em}

\begin{subfigure}{\textwidth}
\centering
\includegraphics[width=0.99\textwidth]{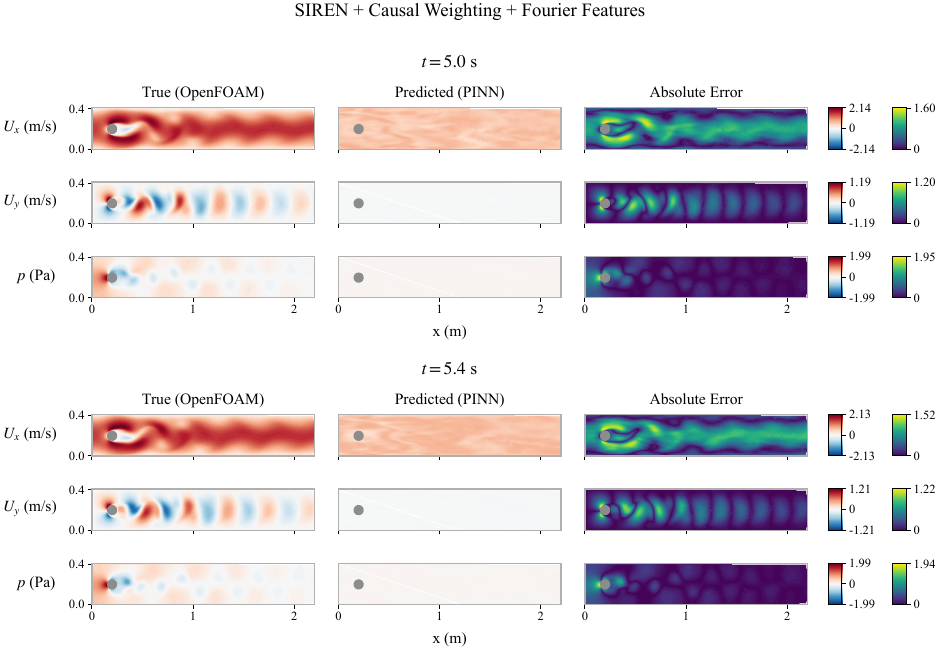}
\caption{SIREN + Causal Weighting + Fourier Features. The predicted fields collapse to a near-uniform, near-zero output across the entire domain, consistent with the output-layer collapse mechanism discussed in Section~\ref{sec:arch-incompat}.}
\label{fig:apdx-ff}
\end{subfigure}
\end{figure}
\newpage

\end{document}